\documentclass[pdflatex,sn-mathphys-num]{sn-jnl}

\usepackage{graphicx}%
\usepackage{multirow}%
\usepackage{amsmath,amssymb,amsfonts}%
\usepackage{amsthm}%
\usepackage{mathrsfs}%
\usepackage[title]{appendix}%
\usepackage{xcolor}%
\usepackage{textcomp}%
\usepackage{manyfoot}%
\usepackage{booktabs}%
\usepackage{algorithm}%
\usepackage{algorithmicx}%
\usepackage{algpseudocode}%
\usepackage{listings}%
\usepackage{subfig}

\theoremstyle{thmstyleone}%
\theoremstyle{thmstyletwo}%

\theoremstyle{thmstylethree}%

\begin{document}

\title[Article Title]{Marker-Based Extrinsic Calibration Method for Accurate Multi-Camera 3D Reconstruction}

\author*[1]{\fnm{Nahuel} \sur{Garcia-D'Urso}}\email{nahuel.garcia@ua.es}

\author[1]{\fnm{Bernabe} \sur{Sanchez-Sos}}\email{bernabe.sanchez@ua.es}
\equalcont{These authors contributed equally to this work.}

\author[1]{\fnm{Jorge} \sur{Azorin-Lopez}}\email{jazorin@ua.es}
\equalcont{These authors contributed equally to this work.}

\author[1]{\fnm{Andres} \sur{Fuster-Guillo}}\email{fuster@ua.es}
\equalcont{These authors contributed equally to this work.}

\author[1]{\fnm{Antonio} \sur{Macia-Lillo}}\email{a.macia@ua.es}
\equalcont{These authors contributed equally to this work.}

\author[1]{\fnm{Higinio} \sur{Mora-Mora}}\email{hmora@ua.es}
\equalcont{These authors contributed equally to this work.}

\affil*[1]{\orgdiv{Department of Computer Science and Technology (DTIC)}, \orgname{University of Alicante}, \orgaddress{\street{San Vicente del Raspeig}, \city{Alicante}, \postcode{03690}, \country{Spain}}}


\abstract{Accurate 3D reconstruction using multi-camera RGB-D systems critically depends on precise extrinsic calibration to achieve proper alignment between captured views. In this paper, we introduce an iterative extrinsic calibration method that leverages the geometric constraints provided by a three-dimensional marker to significantly improve calibration accuracy. Our proposed approach systematically segments and refines marker planes through clustering, regression analysis, and iterative reassignment techniques, ensuring robust geometric correspondence across camera views. We validate our method comprehensively in both controlled environments and practical real-world settings within the Tech4Diet project, aimed at modeling the physical progression of patients undergoing nutritional treatments. Experimental results demonstrate substantial reductions in alignment errors, facilitating accurate and reliable 3D reconstructions.}

\keywords{Multi-camera calibration, RGB-D, 3D Reconstruction}

\maketitle

\section{Introduction}\label{sec1}
In recent years, the demand for high-precision 3D human models has significantly grown across diverse industries, including fashion and garment design, anthropometric measurement extraction, virtual and augmented reality, and gaming. This increasing interest has propelled advancements in multi-sensor technologies, offering enhanced precision and versatility in capturing detailed 3D data.

Among various technologies employed for generating accurate 3D models, RGB-D cameras have emerged as highly effective solutions due to their ability to combine standard RGB imaging with depth sensing. These cameras offer cost-effective, robust, and detailed 3D representations, making them popular in numerous applications. Different RGB-D camera technologies such as structured light, time-of-flight, and stereoscopic systems have distinct advantages tailored to specific application requirements.

Despite their widespread adoption, RGB-D cameras still encounter substantial challenges, primarily related to calibration, which significantly impacts the quality of resulting reconstructions. Accurate calibration is crucial in multi-camera systems and involves determining both intrinsic parameters for individual sensors and extrinsic parameters for camera sets. Calibration errors can result in misaligned depth and color data, severely compromising the integrity and precision of the reconstructed models. Numerous calibration techniques, including iterative and marker-based approaches, have been developed to address these issues, yet perfect integration between multiple sensors remains challenging \cite{PARK2022108535,Chaochuan_2020,9660917}.

Beyond calibration, environmental conditions such as lighting variations, occlusions, and sensor noise significantly influence RGB-D camera performance. Each sensor type exhibits particular vulnerabilities; structured light systems struggle in poor or overly intense lighting, whereas time-of-flight sensors face accuracy reductions at longer distances or in dynamic scenarios. These limitations necessitate sophisticated algorithms to enhance robustness and adaptability \cite{10354343,Atsushi,Chaochuan_2020}.

Additionally, multi-camera arrangements inherently introduce complexity, demanding precise synchronization and alignment to accurately reconstruct 3D scenes. Although significant advancements in calibration and multi-view reconstruction have been made, persistent issues such as depth measurement errors and the requirement for extensive post-processing underscore the ongoing need for research that further refines RGB-D system capabilities \cite{9660917,Curto20213DRO}.

Within this context, the Tech4Diet project focuses on generating 4D models (3D plus temporal dimension) of patients undergoing nutritional treatments. The objective is to accurately capture patient body changes across multiple sessions, presenting their physical progression through augmented reality visualization. This visualization serves as a motivational tool, increasing patient engagement, enhancing treatment adherence, reducing dropout rates, and ultimately improving weight-loss outcomes.

In this paper, we propose an iterative, marker-based calibration methodology specifically designed for multi-camera RGB-D systems. Our approach employs a three-dimensional cube-shaped marker that offers key advantages over traditional planar or spherical calibration objects. Thanks to its geometric design, the marker ensures that multiple planes are simultaneously visible from almost any viewpoint, significantly enhancing robustness and simplifying capture requirements. Unlike spherical-based methods, which typically fit a single model and are highly sensitive to ellipse fitting accuracy, our method independently fits three planar models—one per visible face—followed by the enforcement of strict orthogonality constraints between them. This multi-model fitting strategy improves both the local and global calibration robustness, enabling accurate alignment even under suboptimal viewing conditions or partial occlusions. As a result, the proposed method offers a versatile and highly reliable solution for extrinsic calibration across diverse multi-camera configurations. To evaluate our method’s effectiveness, we conducted three primary experiments examining different configurations of marker height and position. Furthermore, we optimized system hyperparameters to enhance calibration accuracy, demonstrating the applicability of our approach through the successful reconstruction of human body 3D models. The implementation of our proposed calibration method and dataset is publicly available at: \hyperlink{https://github.com/Tech4DLab/CalibMarker}{Github}.

The remainder of the paper is structured as follows: Section \ref{RelatedWork} reviews the relevant literature, Section~\ref{sec:method} outlines the materials and methods, including camera setup details, marker design specifics, and the reconstruction methodology. Section~\ref{Experimentacion} describes the experimental setups and discusses the obtained results. Finally, Section~\ref{Conclusion} provides concluding remarks and highlights potential future research directions.

\section{Related Work} \label{RelatedWork}

Recent advancements in RGB-D camera technology have greatly enhanced the capabilities of multi-camera systems, enabling a diverse range of applications such as autonomous driving, indoor navigation, and 3D reconstruction. A crucial step in achieving precise multi-view fusion and accurate 3D reconstructions is extrinsic calibration, which defines the spatial relationships among the cameras. Traditionally, extrinsic calibration has relied on planar targets such as checkerboards or calibration boards. While effective in controlled environments, these methods often prove inadequate in complex multi-camera configurations. Numerous methods have been proposed to tackle this challenge. Gao et al. \cite{GAO2025115561} presented a methodology utilizing sparse point clouds derived from a 3D object with a texture-rich surface to simultaneously estimate intrinsic and extrinsic camera parameters. Their approach reduces the calibration complexity by employing graph-based optimization techniques. Dai et al. \cite{s24165228} further addressed the calibration of non-overlapping multi-camera systems, a particularly challenging scenario, by generating independent sparse 3D maps and subsequently determining optimal feature correspondences to estimate extrinsic parameters. Shen et al. \cite{Shen20112356} built a non-planar calibration object with sphere and formed a visual sensor network with multiple cameras to calibrate multiple cameras. However, the precision of the method based on sphere is greatly influenced by the precision of ellipse fitting.

Mehmandar et al. \cite{mehmandar2024neuralrealtimerecalibrationinfrared} presented a neural network-based recalibration framework for real-time adjustment of infrared multi-camera systems. Their approach dynamically adjusts camera poses using a differentiable projection model, thereby enhancing robustness against environmental perturbations. Concurrently, Dexheimer et al. \cite{9691878} proposed an information-theoretic approach focused on online extrinsic calibration. Their methodology minimizes entropy and selects informative keyframes to achieve robust and computationally efficient calibration in dynamic settings.

Addressing dynamic and complex environments, recent research has integrated calibration processes within simultaneous localization and mapping (SLAM) frameworks. Dynamic object handling and robust extrinsic calibration are emphasized to cope with real-world complexities. This integrated calibration and SLAM approach facilitates robust extrinsic parameter estimation even in environments with significant scene dynamics and occlusions \cite{10143832}.

Multi-sensor fusion involving cameras and LiDAR sensors has also seen notable advancements. Grammatikopoulos et al. \cite{s22155576} proposed a straightforward yet robust spatiotemporal calibration method for camera-LiDAR systems, utilizing a simple calibration target to effectively align spatial and temporal measurements. This method greatly improves the accuracy and reliability of sensor data fusion in mobile mapping scenarios.

Moreover, recent methods specifically tailored for RGB-D camera setups have emerged. Shin et al. \cite{shin2024pelicaltargetlessextrinsiccalibration} proposed a targetless calibration method called "PeLiCal," leveraging line features to robustly calibrate systems with limited camera overlap. Curto et al. \cite{Curto20213DRO} described a multi-camera RGB-D setup optimized for reconstructing deformable objects, using bright-spot trajectories for extrinsic calibration. Additionally, He et al. \cite{10418976} introduced an automatic extrinsic calibration approach for infrastructure RGB-D networks with small fields of view, utilizing a moving checkerboard and pose graph optimization to enhance robustness.

In summary, while most recent methods for extrinsic calibration in multi-camera systems impose strict requirements, such as reliance on specific RGB-D sensor types, rigid camera setups, or highly controlled marker designs, our proposed approach offers greater flexibility. Although it employs a dedicated three-dimensional cube-shaped marker to exploit geometric constraints, it remains sensor-agnostic and adaptable to arbitrary multi-camera configurations. The marker's structure ensures that multiple faces are visible from almost any viewpoint, overcoming visibility limitations typically associated with planar or spherical targets. Moreover, rather than fitting a single model, as in sphere-based methods that are sensitive to ellipse fitting errors, our method independently fits three planes—one for each visible face—and enforces orthogonality constraints between them. This multi-model fitting strategy significantly enhances calibration robustness and accuracy. The only prerequisite for our method is ensuring overlapping views between cameras, enabling robust and scalable calibration in a wide range of practical deployment scenarios.

\begin{figure}[h]
\centering
\includegraphics[width=8cm]{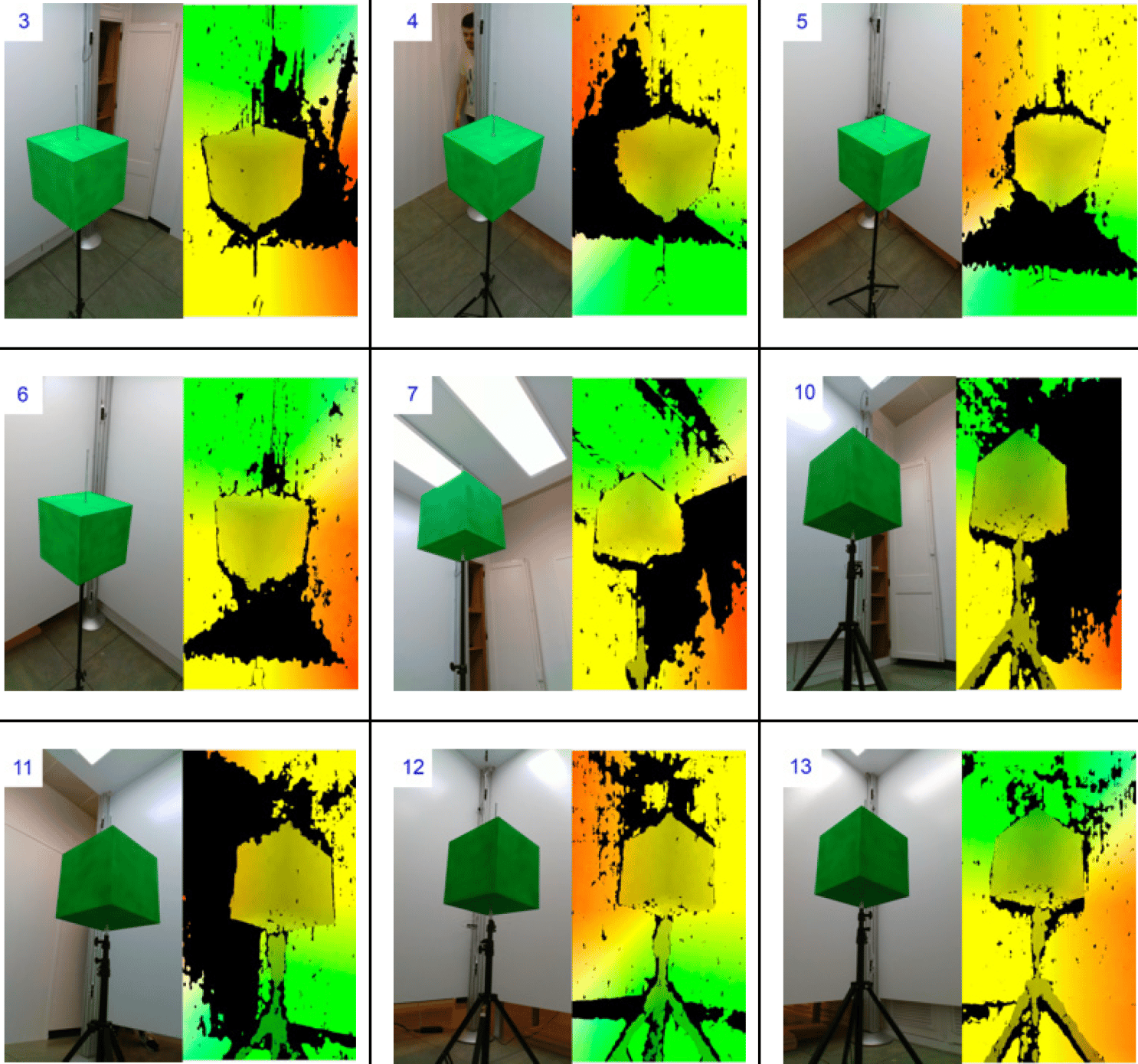}
\caption{Example of RGB and depth images of the cube used for calibration.}\label{fig1}
\end{figure}

\section{Materials and Methods} \label{sec:method}

\subsection{Notations and Setting}

The multi-camera system consists of \(m\) cameras organized in a matrix, named as \(C = \{C_{1}^{1,1}, C_{2}^{1,2}, \dots, C_{m}^{i,j}\}\), where \(i\) and \(j\) indicate respectively the row and column in the camera system. Each camera \(C_{m}^{i,j}\) captures a 3D point cloud \(P_{m}^{i,j}\) and an RGB image \(I_{m}^{i,j}\). The point cloud consists of a set of points \(\{p_{1}, p_{2}, \dots, p_{n}\}\) in Euclidean space, each point being \(p_{n} = [x_{n}, y_{n}, z_{n}]\). A plane \(\pi\) is defined as:
\begin{equation} \label{eq:planeDist}
\pi: \mathbf{n}^\top \mathbf{x} + d = 0,
\end{equation}
where \( \mathbf{n} \in \mathbb{R}^{3} \) is the unit normal vector, \(d \in \mathbb{R}\) is the distance to the origin, and \(\mathbf{x} \in \mathbb{R}^{3}\) represents any point on the plane. For calibration, we use a cube consisting of six planar surfaces \(\{\pi_{1}, \pi_{2}, \dots, \pi_{6}\}\), which satisfies the following geometric constraints:

\begin{itemize}
\item \textbf{Orthogonality:} The normal vectors of the planes satisfy \(\mathbf{n}_{i}^\top \mathbf{n}_{j} = 0\), \(\forall i \neq j\).
\item \textbf{Constant edge length:} All edges of the cube have a known length.
\item \textbf{Visibility constraint:} A camera observes at most three planes of the cube simultaneously.
\end{itemize}

\subsection{Multi-Camera System Calibration}

In this section, we present the proposed methodology for obtaining the transformation matrices (extrinsic calibration) of an RGB-D camera system. The method is divided into five main stages. Firstly, in Section~\ref{segmentacion}, color and depth images are processed. Then, in Section~\ref{fe}, the marker planes are extracted, following an adaptation of the approach presented in~\cite{SAVALCALVO2015572,AzorinLopez2021}. Finally, in Section~\ref{Calibracion}, the calibration matrices that align the different cameras are estimated. The marker used is a cube, as its known geometry allows for the exploitation of robust geometric constraints, such as orthogonality between planes and partial visibility from each camera. Each camera must simultaneously visualize three faces of the cube for the method to function correctly, as shown in Figure~\ref{fig1}.

\begin{figure}[H]
\centering
\includegraphics[width=13cm]{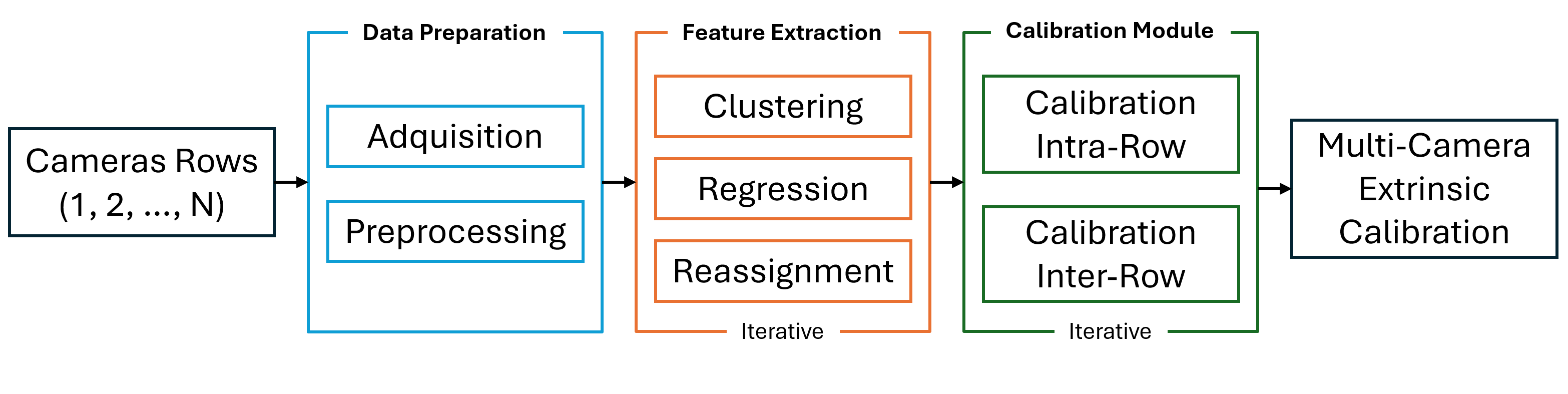}
\caption{Overview of the proposed calibration pipeline. The method is structured into three main modules: Data Preparation, Feature Extraction, and Calibration. Each module includes specific processing stages, starting from RGB-D image acquisition and preprocessing, followed by iterative geometric extraction of cube faces, and ending with intra and inter-row calibration steps. The final output is the extrinsic calibration of the multi-camera system. \label{figPipe}}
\end{figure}

\subsection{Data Preparation} \label{segmentacion}

This stage aims to extract the region of interest from RGB and depth images captured by each camera. It is assumed that all cameras have been previously calibrated intrinsically. Before initiating this process, it is necessary to capture images of the calibration cube at various heights and positions, ensuring at least one capture exists in which cameras from different rows simultaneously observe the cube. Given that the calibration cube is green, the RGB images are first converted to the HSV color space to facilitate effective color-based segmentation. A binary mask is subsequently created by applying predefined thresholds to the hue, saturation, and depth channels. Using region detection, the largest connected region, corresponding to the marker, is identified and segmented in both RGB and depth images. From this segmentation, a filtered 3D point cloud is obtained, containing only the cube region to be used in the subsequent stage.

\subsection{Feature Extraction} \label{fe}

This stage focuses on extracting geometric information from the segmented point cloud \(P_{m}^{i,j}\) to identify the visible faces of the cube and estimate their corresponding planes. The process comprises three main steps clustering, regression, and reassignment which are applied iteratively.

The clustering stage involves identifying and grouping 3D points from the segmented point cloud \(P_{m}^{i,j}\) into distinct flat regions that correspond to the visible faces of the cube. The objective of the clustering stage is to segment the point cloud into \(N\) groups, where \(N\) corresponds to the number of visible planes of the cube (three per camera view). This process uses both the 3D coordinates of the points \(\mathbf{p} = [x, y, z]\) and their normal vectors \(\mathbf{n} = [n_x, n_y, n_z]\). Each point's normal is computed by averaging the normals of its neighboring points.

Points and their normals are grouped using the K-means algorithm, where each resulting cluster represents a plane \(\pi\) corresponding to one of the visible faces of the cube. Clusters with insufficient point density are discarded to improve robustness. Additionally, the angular consistency of each group is verified by comparing the angles between individual normal vectors and the median group's normal vector. Points with large angular deviations are excluded from the cluster.

To ensure that each group represents a single plane \(\pi\), groups with similar normal vectors or those that are spatially close are merged. This merging process uses the orthogonality constraint. Since the object to be reconstructed is known (a cube), it is understood that the angle between different planes must always be 90 degrees. Finally, for each obtained cluster, its centroid \(\mathbf{c}\) and normal vector \(\mathbf{n}\) are calculated.

The regression stage refines the planes obtained during clustering by fitting each group of points to a mathematical representation of a cube (identical to the real one). The aim of the regression is to estimate the parameters \(\mathbf{n}\) and \(d\) of a plane that minimize the error, that is, the distance between the points of the group and the plane of a cube model.

For a group of points \(\mathbf{p}_1, \mathbf{p}_2, \mathbf{p}_3\) associated with a plane \(\pi\), the regression process begins by calculating the centroid of the points:
\begin{equation} \label{eq:centroid}
\mathbf{C} = \frac{1}{N} \sum_{i=1}^N \mathbf{p}_i.
\end{equation}

Then, the normal vector \(\mathbf{n}\) is calculated using principal component analysis (PCA). The ultimate goal of this step is to verify that the obtained clusters meet the necessary orthogonality constraints to create a cube. That is, there is a 90-degree angle between them, thus enabling the generation of a cube. 

The reassignment stage seeks to validate the belonging of a point to a plane. For each point \(\mathbf{p}_i\) in the segmented point cloud \(P{m}^{i,j}\), its distance to all candidate planes \(\{\pi_1, \pi_2, \pi_3\}\) is calculated using the plane equation \ref{eq:planeDist}. The point is reassigned to the plane \(\pi_k\) that minimizes this distance, ensuring that each point is associated with the closest plane. Then, the similarity of the point's normal with the normal of the assigned plane is checked. To calculate the point's normal, the average between its normal and those of adjacent points is computed.

In the next step, the angular consistency between the point's normal vector \(\mathbf{n}_{\mathbf{p}_i}\) and the plane's normal vector \(\mathbf{n}_{\pi_i}\) is evaluated. Points with angular deviations greater than a predefined threshold are excluded. Only points that meet the angular constraint are kept in the reassigned groups. This primarily occurs on the edges, where a point may be close to a plane but that plane has a different normal.

\begin{figure}[h]
\centering
    \includegraphics[width=8cm]{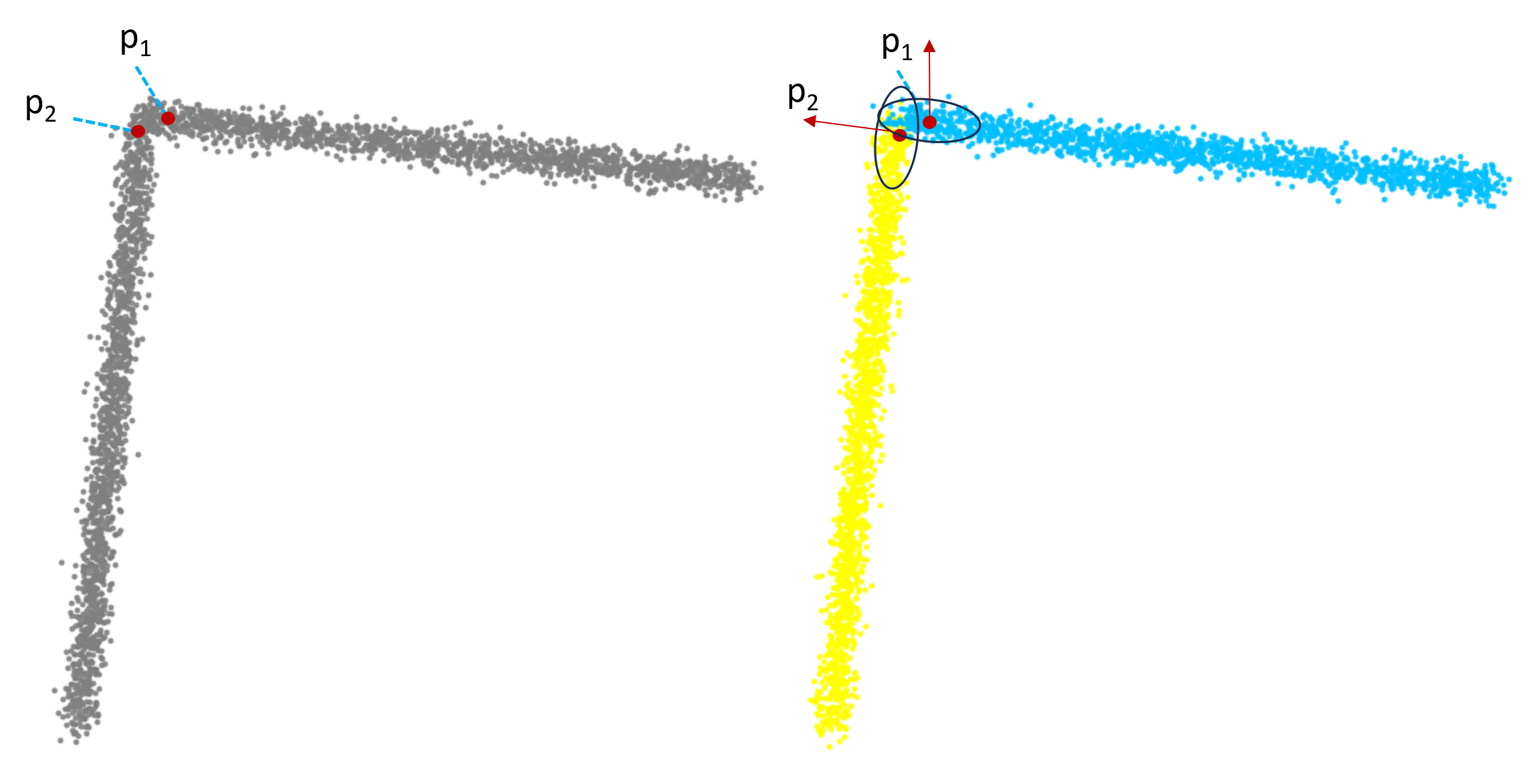}
    \caption{Example of two points erroneously assigned to the same plane.}
\label{fig:cluster-comparison2}
\end{figure}

\begin{figure}[h]
\centering
    \subfloat[\centering]{\includegraphics[width=4cm]{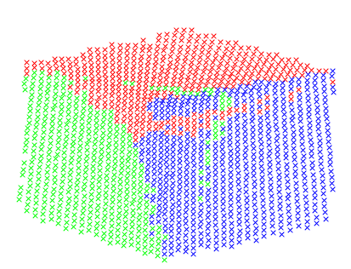}}
    \hfil
    \subfloat[\centering]{\includegraphics[width=4cm]{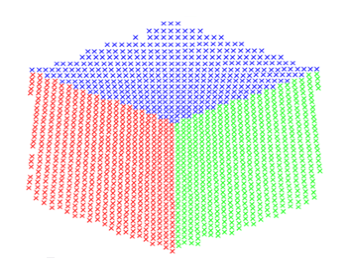}}
\caption{Comparison of cluster assignments in different phases of the algorithm. (a) Cube model with points assigned in wrong clusters at the beginning of the algorithm. (b) The same cube after iterating multiple times over the Cluster, Regression, and Reassignment phases.}
\label{fig:cluster-comparison}
\end{figure}

The reassignment process is carried out iteratively to refine the association between points and planes. After each iteration, the \(n\) and \(d\) parameters of the planes are recalculated using the updated groups, and the reassignment is repeated. The Clustering, Regression, and Reassignment stages are repeated iteratively until all points belong to their corresponding plane. This iterative refinement continues until all points belong to their respective plane. An example of the iterative process of obtaining the planes of the cube can be observed in the figures \ref{fig:cluster-comparison2} and \ref{fig:cluster-comparison}. 

\subsection{Calibration} \label{Calibracion}

The calibration step computes the transformation matrices that align the multi-camera system \(C = \{C_{1}^{1,1}, C_{2}^{1,2}, \dots, C_{m}^{i,j}\}\) by estimating the relative positions and orientations of all cameras within a common reference frame. Each transformation matrix \(\mathbf{T}_{m}^{i,j} \in SE(3)\) is a rigid $4 \times 4$ transformation composed of a rotation matrix \(\mathbf{R} \in SO(3)\) and a translation vector \(\mathbf{t} \in \mathbb{R}^3\), which together describe the pose of the camera in space.

To compute these transformations, the calibration algorithm utilizes geometric features extracted from a known marker. Specifically, it focuses on the centroids and normals of the three visible faces of a cube in each camera's view. To enhance robustness against outliers and noise, a traditional RANSAC (Random Sample Consensus) method is employed to eliminate inconsistent observations. Following this, a Procrustes analysis is used to estimate the transformation that best aligns the observed planes from each camera to a chosen reference view. The alignment process is conducted jointly on the centroids and normals to ensure both positional and angular consistency

The algorithm structures the camera system as a graph, where each node corresponds to a camera and edges represent pairs of cameras with simultaneous observations of the calibration marker. Calibration is then performed in two sequential stages.

\begin{enumerate}
    \item \textbf{Intra-row calibration:} For each row of cameras, a reference camera is selected, and its transformation matrix is fixed to the identity, \(T = I\). The other cameras in the same row are then calibrated relative to this reference camera by estimating the rigid transformations needed to align their observed planes.
    \item \textbf{Inter-row calibration:} After each row has been calibrated internally, the rows are connected using overlapping views. For example, cameras in row 2 that share views with cameras in row 1 are used to calculate the relative transformations between the rows. The same procedure is repeated between all rows. This results in a complete set of transformation matrices for all the cameras in the system.
\end{enumerate}

This graph-based strategy allows for a scalable and structured calibration, even in systems with limited overlap between some camera pairs. During each alignment step, an error metric is computed to assess calibration quality. This metric combines two components: the average Euclidean distance between corresponding centroids and the mean angular deviation between their normal vectors. The total alignment error is defined as a weighted sum of both terms:

\begin{equation}
\text{Error} = \alpha \cdot \text{Distance Error} + \beta \cdot \text{Angular Error},
\end{equation}

where \(\alpha\) and \(\beta\) are weighting factors that control the relative importance of spatial and directional accuracy.

Once the iterative process converges, each camera will have an associated transformation matrix \(\mathbf{T}_{m}^{i,j}\). This matrix allows the point cloud from each camera to be expressed in a unified coordinate system defined by the reference camera. This approach ensures spatial consistency across the entire camera system, enabling accurate and coherent 3D reconstruction.


\section{Experimentation} \label{Experimentacion}

\subsection{Experimental Setup} \label{sec:experiments}

Two experiments were conducted to quantitatively validate the proposed method. The first experiment assesses the number of captures necessary to accurately calculate the marker planes. The objective is to determine the optimal number of captures and cube positions required for precise calibration. The second experiment focuses on optimizing the hyperparameters of our method. Specifically, we aim to identify the best parameter settings for reconstructing an object within the capture zone.

To evaluate the quality of the calibration matrix obtained with our method, we use various metrics. The size difference between the reconstructed and ground truth objects is computed. Additionally, we calculate the angle, in degrees, of the reconstructed cube's planes. This is done for each row of cameras and for the entire system as a whole.

Furthermore, we use Wasserstein and Hausdorff distances to evaluate the accuracy of our system in reconstructing human body models.

The Hausdorff distance between two sets of points \(A\) and \(B\) is defined as:

\begin{equation}
d_H(A, B) = \max \left\{ \sup_{a \in A} \inf_{b \in B} d(a, b), \sup_{b \in B} \inf_{a \in A} d(a, b) \right\},
\end{equation}

where \(d(a,b)\) represents the distance between the points \(a \in A\) and \(b \in B\). This metric measures the maximum distance from a point in one set to the closest point in the other set.

The Wasserstein distance quantifies the minimum effort required to transform one distribution \(P\) into another \(Q\). In the context of 3D reconstruction, it evaluates the structural similarity between the reconstructed and reference models.

Additionally, we present qualitative results. For this, we perform the reconstruction of a human body using different calibrations obtained during experimentation. This human body belongs to one of the patients examined in the Tech4Diet project.

\begin{figure}[h]
\centering
\subfloat[\centering]{\includegraphics[width=3.9cm]{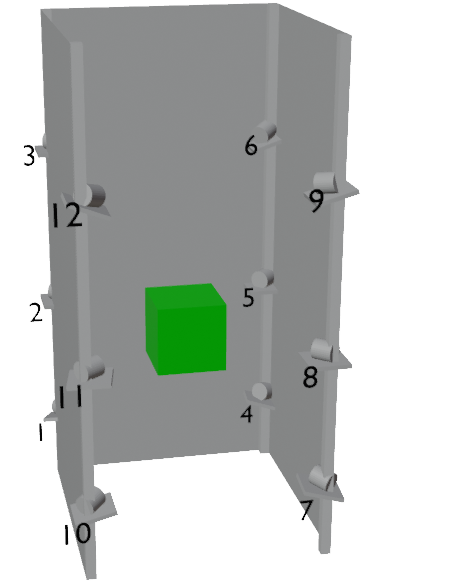}}
\subfloat[\centering]{\includegraphics[width=3.9cm]{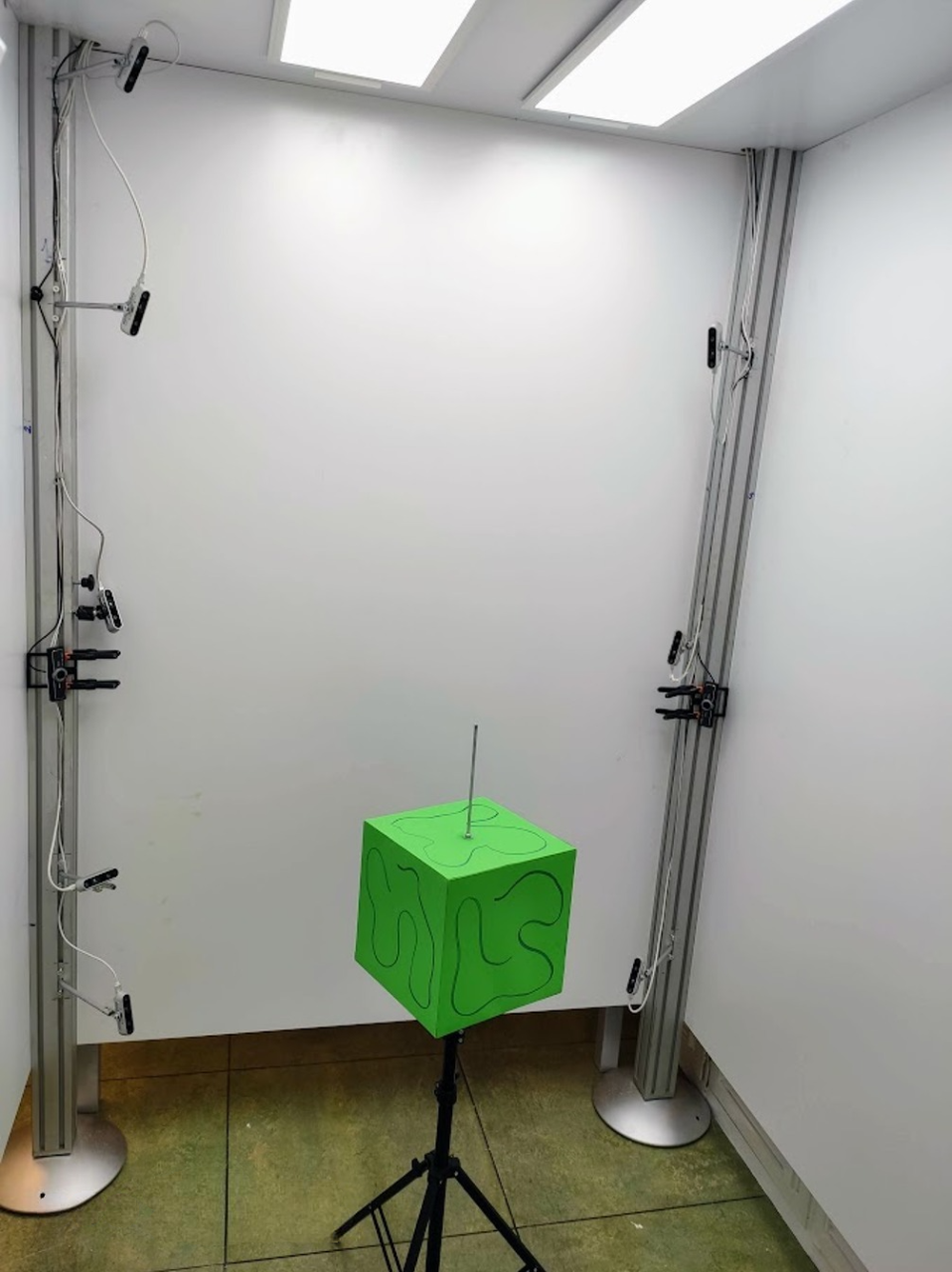}}
\caption{On the left, the 3D simulation of the booth using Blender. On the right, the designed booth prototype. \label{fig:blender}}
\end{figure}

To determine the optimal camera configuration, a 3D simulation was carried out using Blender, figure \ref{fig:blender}. This setup aimed to maximize coverage of the scanned volume, minimize occlusions, and ensure sufficient overlap between adjacent views for accurate reconstruction. The final synthetic configuration consists of 12 RGB cameras. The Intel RealSense camera was selected due to its low cost, compact design, and wide field of view.

\subsection{Quantitative Results}
\begin{table*}[tbp]
\centering
\caption{Mean positional (mm) and angular (degrees) errors for different heights and captures.} \label{tab:errores}
\resizebox{\textwidth}{!}{%
\begin{tabular}{cc cccccc}
\toprule
\multirow{2}{*}{\textbf{\#Captures}} & \multirow{2}{*}{\textbf{\#Heights}} & \multicolumn{5}{c}{\textbf{Errors (mm - degrees)}} & \multirow{2}{*}{\textbf{Total (mm - degrees)}} \\ 
\cmidrule(lr){3-7} 
& & Row 1 & Row 2 & Row 3 & Row 2-1 & Row 3-2 & \\ 
\midrule
4  & 3  & 0.0014 - 0.63  & 0.0018 - 1.51  & 0.0013 - 3.11  & 0.0034 - 0.65  & 0.0035 - 4.04  & 0.0012 - 1.62 \\ 
4  & 6  & 0.0015 - 0.45  & 0.0019 - 1.19  & 0.0012 - 1.27  & 0.0038 - 0.60  & 0.0030 - 3.10  & \textbf{0.0012 - 0.83} \\ 
4  & 9  & 0.0015 - 0.44  & 0.0128 - 3.13  & 0.0147 - 3.73  & 0.0037 - 0.83  & 0.0553 - 12.64 & 0.0064 - 1.71 \\ 
8  & 3  & 0.0013 - 0.44  & 0.0039 - 3.08  & 0.0041 - 3.39  & 0.0031 - 0.49  & 0.0138 - 12.28 & 0.0022 - 1.63 \\ 
8  & 6  & 0.0015 - 0.42  & 0.0041 - 2.23  & 0.0042 - 2.37  & 0.0034 - 0.51  & 0.0142 - 8.31  & 0.0023 - 1.22 \\ 
8  & 9  & 0.0015 - 0.37  & 0.0098 - 3.14  & 0.0112 - 3.69  & 0.0034 - 0.67  & 0.0409 - 12.80 & 0.0051 - 1.68 \\ 
\bottomrule
\end{tabular}%
}
\end{table*}

\begin{figure}[h]
\centering
\includegraphics[width=10cm]{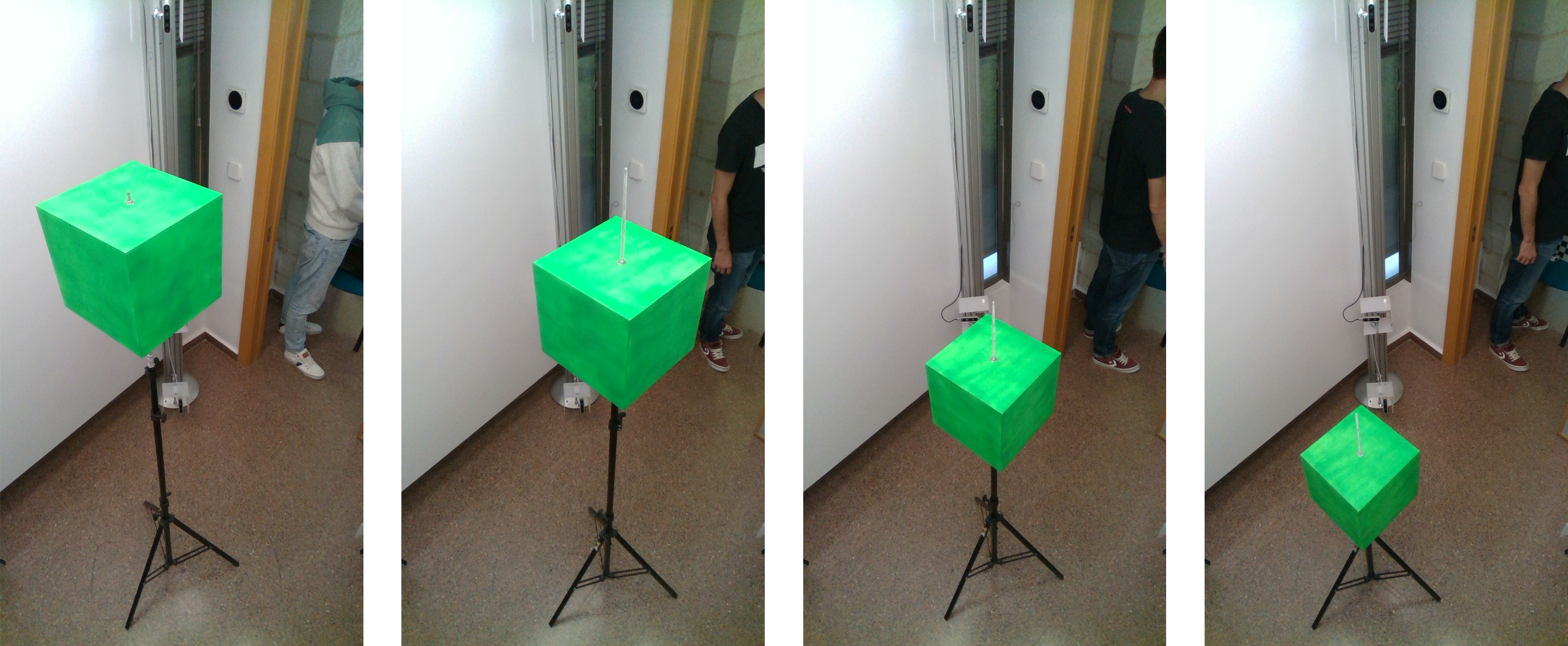}
\caption{View of the cube at different heights for a camera in the system during calibration. \label{fig:cube1}}
\end{figure}

In Table \ref{tab:errores}, the results obtained by reconstructing the cube in different scenarios can be observed. The first column shows the number of captures made for each position of the cube, while the second shows the different heights of the cube. Given that the system has three rows of cameras, in the simplest case, rows 1 and 4 of the table, we find three heights of the cube.
An example of placing the cube at different heights can be seen in the figure \ref{fig:cube1}. In the columns with the name Row 1,2, and 3, the results of the cube reconstruction for each row individually can be seen. The columns Row 2-1 and Row 3-2 refer to the results of reconstructing the cube using both rows at the same time

In terms of the results, it can be observed that the configuration of 6 heights of the cube with 4 captures for each cube obtained the best results both in terms of reconstruction of the cube in millimeters and degrees obtained in the planes of the cube.

Using this configuration, we optimized the hyperparameters of our method by conducting a grid search involving 322 different configurations. The mean absolute error (MAE) was used as the loss metric to select the best hyperparameter configuration. In Table \ref{tab:hyper}, the hyperparameters used can be observed. This optimization led to improved reconstruction accuracy. The error in millimeters has remained at the same value but the angular errors were reduced by 50\%, from 0.83 to 0.41 degrees in the reconstruction of the cube using the transformation matrices obtained with calibration.

\begin{table}[h]
\centering
\caption{Values of the hyperparameters used to obtain optimal calibration.}
\label{tab:hyper}
\small
\begin{tabular}{ll}
\toprule
\textbf{Parameter}                                                                                         & \textbf{Values}         \\
\midrule
\multirow{2}{*}{\begin{tabular}[c]{@{}l@{}}\#Minimum captures for \\ fitting the cube model\end{tabular}} & \multirow{2}{*}{2, 3, 5} \\ \\
Maximum iterations                                                                                        & 25, 50, 100              \\
Distance threshold                                                                                        & 0.003, 0.006, 0.01       \\
Angular threshold                                                                                         & 0.3, 0.6, 1              \\
Distance threshold between rows                                                                           & 0.1, 0.001               \\
Angular threshold between rows                                                                            & 5, 3, 2                  \\
Considering normals                                                                                       & True, False              \\
\bottomrule
\end{tabular}
\end{table}

In Table \ref{tab:errores3}, the results of calculating the Hausdorff and Wasserstein distances in 3D models of the human body reconstructed using different calibration settings can be observed. To perform these calculations, the optimized calibration that obtained the best results in the reconstruction of the cube was taken as ground truth. It can be seen how the other configurations, with the exception of the configuration of 3 heights and 4 captures, perform significantly worse. This is even more evident in the images shown in the qualitative results \ref{fig1:2}.

\begin{table}[h]
\centering
\caption{Hausdorff and Wasserstein distances obtained from the optimized model and the non-optimized ones.}
\label{tab:errores3}
\small
\begin{tabular}{ccc}
\toprule
\multirow{2}{*}{\textbf{\begin{tabular}[c]{@{}c@{}}Configuration\\ \#Captures - \#Heights\end{tabular}}} & \multicolumn{2}{c}{\textbf{Distances}} \\
 & \textbf{Hausdorff} & \textbf{Wasserstein} \\
\midrule
4 - 3 & 0.11 & 0.043 \\
4 - 6 & 0.11 & 0.040 \\
4 - 9 & 0.49 & 0.101 \\
8 - 3 & 0.37 & 0.086 \\
8 - 6 & 0.27 & 0.069 \\
8 - 9 & 0.42 & 0.087 \\
\bottomrule
\end{tabular}
\end{table}

\subsection{Qualitative Results}

\begin{figure}[H]
\centering
\includegraphics[width=8cm]{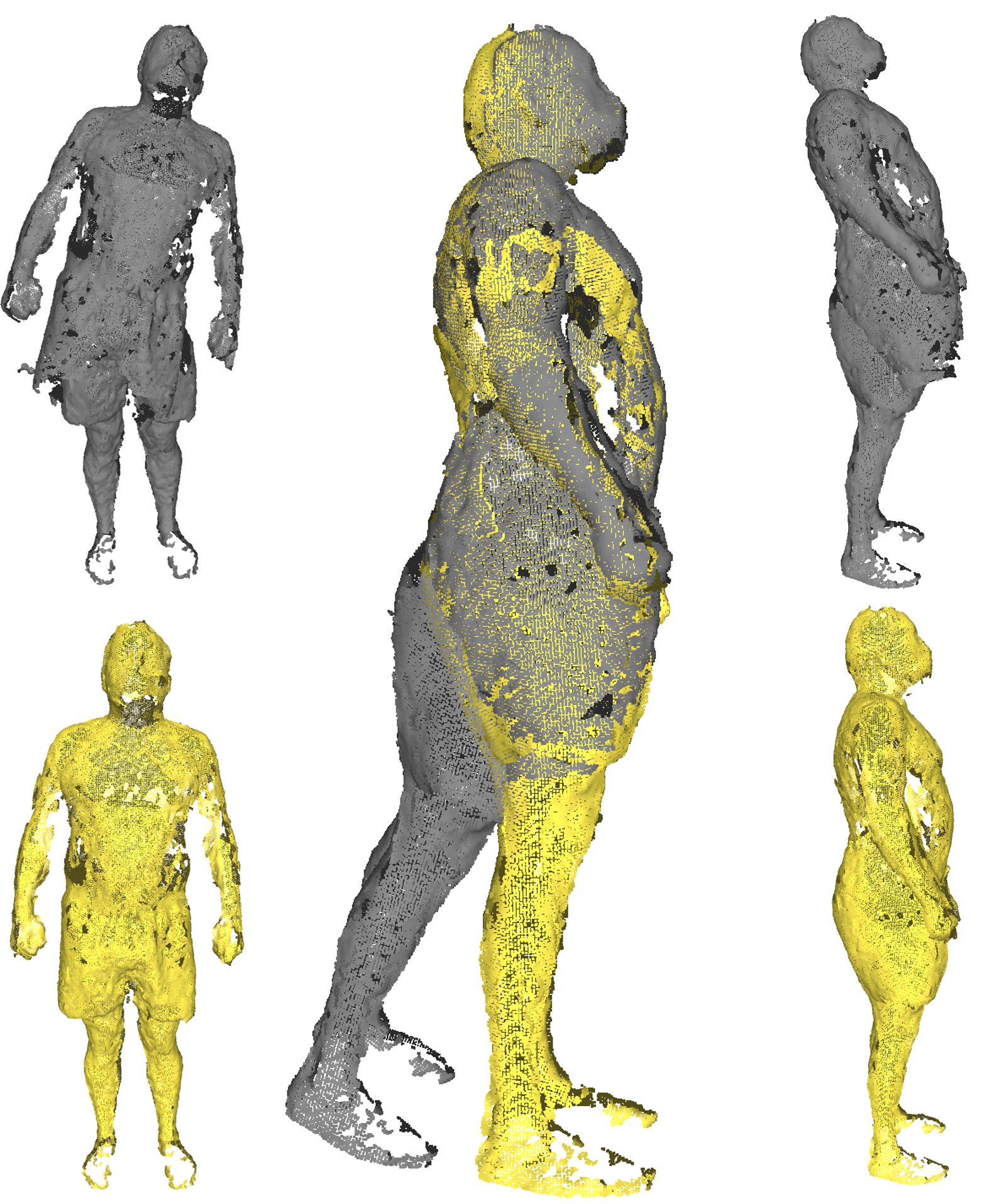}
\caption{Comparison of point clouds registered using the optimized calibration (yellow) and non-optimized (gray). \label{fig1:2}}
\end{figure}

For the qualitative results in this work, we have reconstructed a human body using the calibration matrices obtained during quantitative experimentation. In Figure \ref{fig1:2}, the difference between the bodies reconstructed from the optimized and non-optimized calibration can be seen. In the central image, it is observed how from the optimized calibration the leg area aligns correctly with the rest of the body. While in the non-optimized one, a backward displacement of the lower limbs is observed due to calibration errors.

\begin{figure}[H]
\centering
\includegraphics[width=8cm]{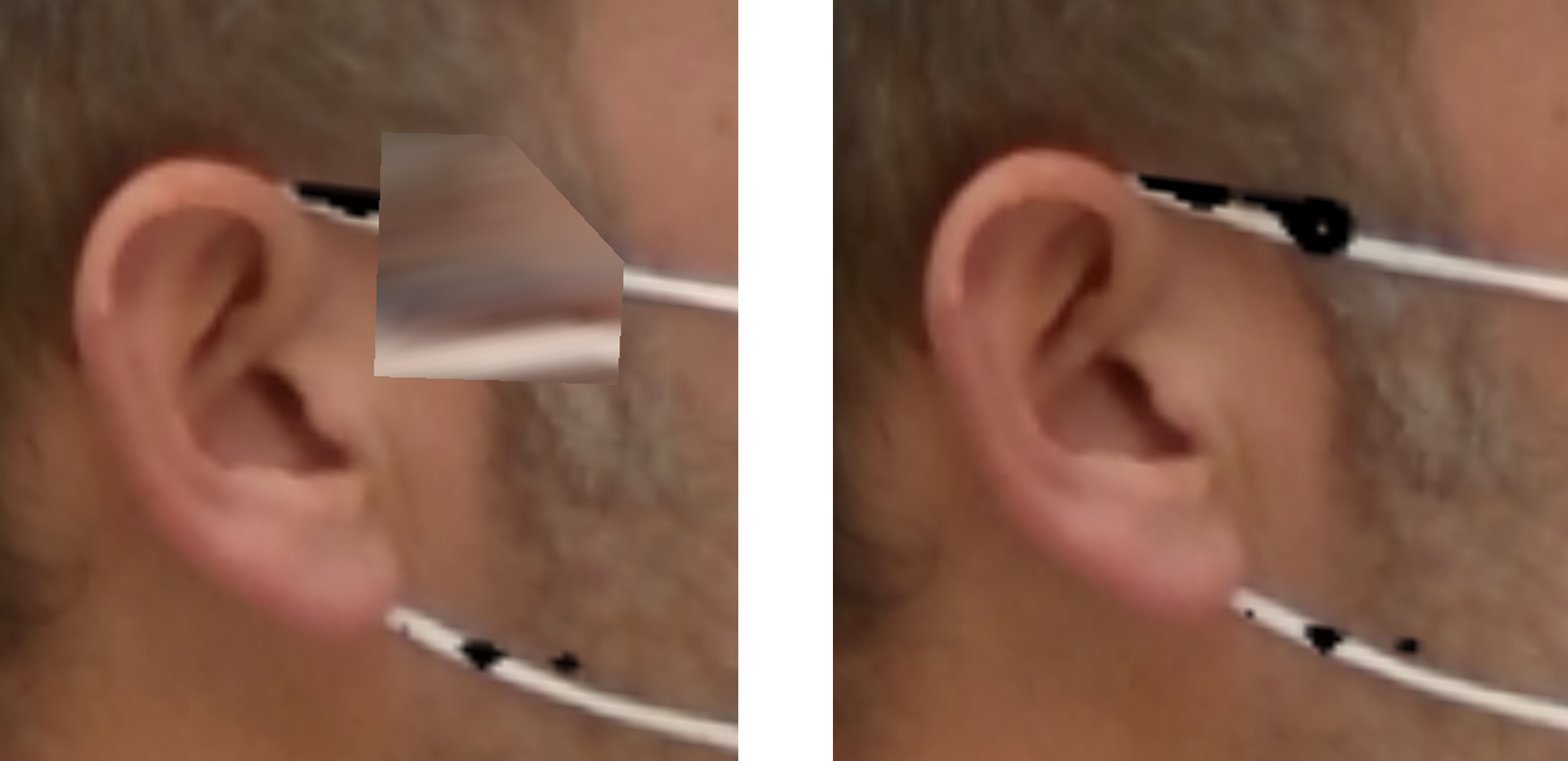}
\caption{Comparison of the texture generated from the model reconstructed with the non-optimized calibration (left) with the optimized one (right) \label{fig3}}
\end{figure}

But having an optimal calibration not only improves point cloud registration but also enhances mesh generation and texture projection. As shown in Figure \ref{fig3}, the use of optimal calibration leads to cleaner texture projection with fewer visual artifacts.

\begin{figure}[H]
\centering
\includegraphics[width=10cm]{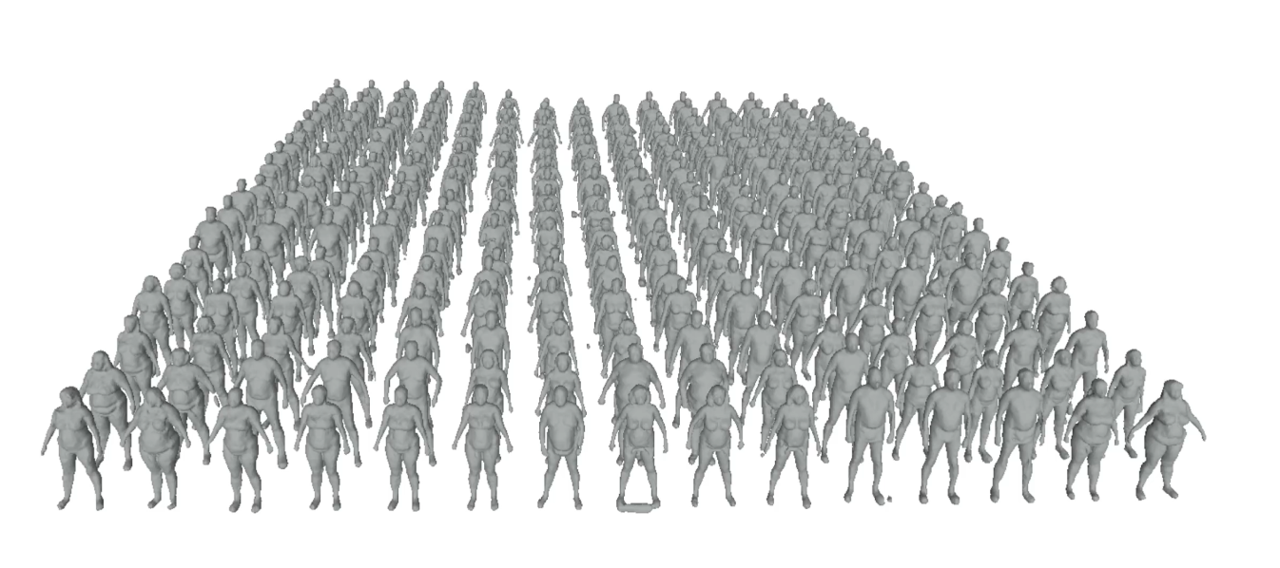}
\caption{Examples of 3D human body reconstructions obtained using the proposed calibration method within the Tech4Diet project. \label{fig:dataset1}}
\end{figure}

\begin{figure}[H]
\centering
\includegraphics[width=8cm]{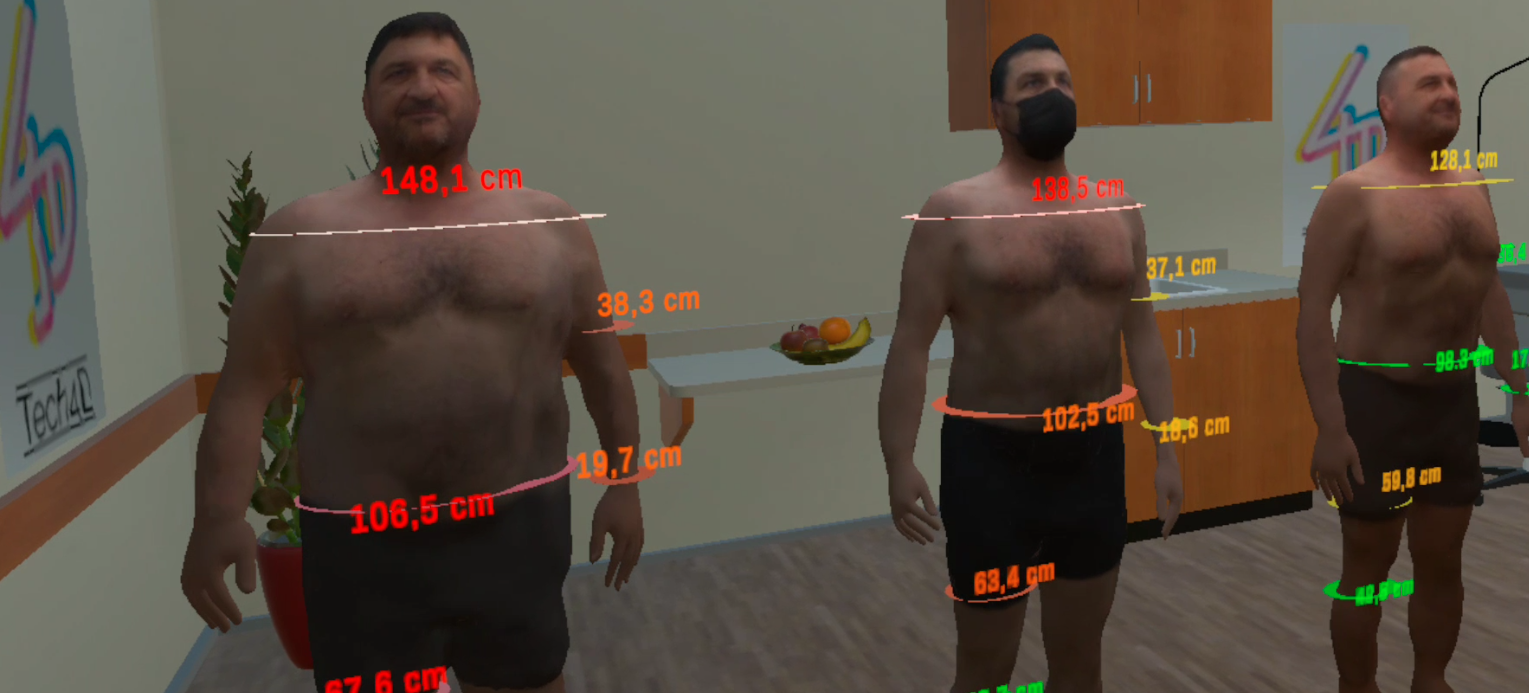}
\caption{Visualization of a patient's reconstructed body models through the Virtual Reality application, used as a supportive tool during nutritional interventions. \label{fig:vr}}
\end{figure}

The proposed calibration method has been employed to reconstruct more than 300 human body models as part of the Tech4Diet project. Several examples of these reconstructions, achieved using the optimized calibration, are shown in Figure \ref{fig:dataset1}. Patients visualize their reconstructed body models via a Virtual Reality application (Figure \ref{fig:vr}), which serves as an assistive tool to enhance adherence and engagement with their nutritional treatment.

\section{Conclusion} \label{Conclusion}

In this work, a marker-based multi-camera calibration method has been presented. The experiments performed demonstrate that the proposed method enhances the accuracy of 3D reconstruction by minimizing positional and angular errors in camera alignment. In particular, it has been observed that an optimal configuration of six marker heights with four captures per position produces the best results, reducing angular error by 50\% after hyperparameter optimization.

A key innovation of the proposed approach is the use of a three-dimensional cube-shaped marker that ensures the visibility of multiple planes from a wide range of viewpoints, overcoming traditional limitations of planar or spherical markers. Unlike methods based on a single model fitting, such as those using spherical targets prone to ellipse fitting errors, our method fits three independent planes—one per visible face—and applies strict orthogonality constraints between them. This multi-model fitting strategy significantly improves calibration robustness and enables highly accurate multi-camera alignment even under partial occlusions or suboptimal conditions.

Beyond this specific application, the developed methodology is extensible to other domains that require precise volumetric capture, such as body scanning in health and sports, avatar generation for virtual reality, film production, or ergonomic monitoring in industrial environments.

While the proposed approach has demonstrated promising results, several open challenges remain. Future work will focus on extending the method to other geometric configurations and validating its accuracy through comparisons between real-world anthropometric measurements and those obtained from the calibrated 3D reconstructions.

\section*{Acknowledgments}
This work has been funded by the Spanish State Research Agency (AEI) through the grant PID2023-149562OB-I00, awarded by the MCIN/AEI/10.13039/501100011033, as well as by the consolidated group project CIAICO/2022/132 "AI4Health", financed by the Government of the Valencian Community.

\bibliography{sn-article}

\end{document}